# WSCIF: A Weakly-Supervised Color Intelligence Framework for Tactical Anomaly Detection in Surveillance Keyframes

Wei Meng

Dhurakij Pundit University, Thailand
PCBM,Sasin School of Management of Chulalongkorn University
Fellow, Royal Anthropological Institute,UK (RAI)

Email: wei.men@dpu.ac.th

## Abstract

The deployment of traditional deep learning models in high-risk security tasks in an unlabeled, data-non-exploitable video intelligence environment faces significant challenges. In this paper, we propose a lightweight anomaly detection framework based on color features for surveillance video clips in a high sensitivity tactical mission, aiming to quickly identify and interpret potential threat events under resource-constrained and data-sensitive conditions. The method fuses unsupervised KMeans clustering with RGB channel histogram modeling to achieve composite detection of structural anomalies and color mutation signals in key frames. The experiment takes an operation surveillance video occurring in an African country as a research sample, and successfully identifies multiple highly anomalous frames related to high-energy light sources, target presence, and reflective interference under the condition of no access to the original data. The results show that this method can be effectively used for tactical assassination warning, suspicious object screening and environmental drastic change monitoring with strong deployability and tactical interpretation value. The study emphasizes the importance of color features as low semantic battlefield signal carriers, and its battlefield intelligent perception capability will be further extended by combining graph neural networks and temporal modeling in the future.

**Keywords:** weakly supervised learning; anomaly detection; color clustering; tactical video analysis

## Chapter I. Introduction

In modern, highly asymmetric conflict environments, tactical intelligence acquisition increasingly relies on fast, low-cost and robust video surveillance systems. However, realistically deployed video data often faces multiple stringent constraints: source anonymity, missing context, unavailable semantic tags, and even destruction after acquisition, leaving only fragmented clips or limited keyframe sequences (Sabokrou et al., 2016). In such a context of information gaps and resource constraints, deep semantic models (e.g., YOLO, CLIP, etc.) relying on a large amount of labeled data and complex computational resources are not only difficult to be deployed, but also not even feasible at the tactical frontline (Ren et al., 2015). Therefore, the development of a set of video intelligence analytics frameworks that can be adapted to "data-constrained", "semantically unlabeled", and "front-end runnable" conditions has become a key requirement for the construction of tactical-level AI systems (Lin et al., 2015). key claim (Lin et al., 2023).

As the most primitive, cheapest and computationally friendly visual feature in images, color information has long played a role in traditional image processing, especially in target recognition, environment classification and anomaly detection tasks with stable performance (Bian & Tao, 2011). Although deep learning-based visual models have made significant progress in high-semantic scenarios in recent years, in real-world tactical applications where semantics are unavailable, bandwidth is limited, or the task is sensitive, color features still show their unique advantages: no training is required, they are interpretable, they have minimal device requirements, and they are particularly suitable for deployment in unmanned platforms, edge nodes, and classified environments (Hasan et al., 2016). However, due to the nonlinear complexity and diversity of color distribution, its direct mapping to semantic labels is still difficult, which puts higher requirements on algorithmic modeling (Pang et al., 2021).

In this paper, we propose a Weak-Supervision Color Intelligence Framework (WSCIF) for tactical video keyframe analysis, which aims to solve the problems of "data unshareability", "context unavailability" and "context-unavailability". The WSCIF aims to solve the problem of tactical information recognition in the context of "unsharable data", "unavailability of context" and "missing semantics". The method combines two lightweight analysis paths, namely unsupervised clustering (KMeans) and channel distribution modeling (RGB histogram), to automatically extract structural anomalies and color mutation signals from the original image space, and construct frame-level anomaly recognition logic to achieve rapid screening and structured modeling of potential threats such as scene mutations, light source anomalies, fires, or reflections (Chen et al., 2020).

In this paper, a video clip from the border surveillance system of an African country is selected for empirical analysis, and multiple color feature-driven anomalies are successfully identified without relying on semantic annotations or accessing the complete video content. Experimental results show that the method achieves highly robust anomaly detection and intelligence indication at the key frame level, and possesses a very low deployment threshold and strong adaptive capability for complex tactical scenarios such as edge computing platforms, unmanned nodes, and temporary deployment networks (Xu & Yang, 2024).

The core contributions of this paper include the following three points:

1).proposing a color intelligent recognition framework without semantic tags and complete videos to achieve tactical anomaly detection at the key frame level;

2). constructing a dual-path analysis mechanism of "clustering + channel response", which integrates structural clustering and color statistical features to significantly improve the interpretability and deployment efficiency of anomaly recognition;

3). verify the practicality of this method in tactical counter-terrorism reconnaissance scenarios, and provide theoretical basis and empirical support for building lightweight and unsupervised video intelligent sensing systems.

The next chapters will sequentially introduce the related research progress (Chapter 2), the methodological framework proposed in this paper (Chapter 3), the experimental setup and process (Chapter 4), the empirical analysis results (Chapter 5), the discussion of the applicability and limitations of the method (Chapter 6), and the conclusion and future research direction (Chapter 7).

## Chapter II. Relevant studies

Tactical video analytics has become a key landing direction for AI in recent years in scenarios such as security and defense, reconnaissance surveillance and edge deployment (Xu & Yang, 2024). In unmanned platforms, portable security systems and temporary frontier surveillance architectures deployed in high-risk areas, facing the limitations of video data source anonymity, missing context and unavailability of semantic labels, how to accomplish effective environment recognition and event detection under extremely resource-constrained conditions has become an urgent problem for AI researchers (Sabokrou et al., 2016; Lin et al. , 2023). Existing research focuses on three directions: video anomaly detection methods, color feature modeling and analysis, and adaptation of weakly supervised learning in reconnaissance situations.

### 2.1 Video Anomaly Detection Methods

Current video anomaly detection methods can be broadly classified into two categories: those based on temporal motion modeling and those based on image appearance features (Chandola et al., 2009). The former relies on Optical Flow, background modeling, or convolutional temporal networks (e.g., ConvLSTM), and has strong modeling capabilities for emergent behaviors such as running, aggregation, or conflict (Liu et al., 2018; Hasan et al., 2016). However, in tactical videos with low frame rates, motion blur, or crippled data, its performance degrades significantly and its generalization ability is weak (Ren et al., 2015). The latter relies on the static structure of images with visual anomalies, such as utilizing a family of Convolutional Autoencoders (CAE) or Generative Adversarial Networks (GAN), to identify anomalous events through reconstruction errors, but this class of methods is highly dependent on the scene consistency of the training data, and is usually not applicable to environments where semantic labels are missing or deployments are not trainable (Sultani et al., 2018).

Some studies have begun to focus on the potential of color information in anomalous frame recognition, such as identifying flames, bloodstains, and strongly reflective targets through color histograms (Zhang et al., 2022), but these methods mostly exist as preprocessing modules, and have yet to form a systematic and structured analysis framework (Bian & Tao, 2011). Especially in tactical-level applications, the constraints of real-time and computational efficiency make lightweight unsupervised anomaly detection methods more promising for application.

**2.2 Color Features and Video Analysis**

Color is the most basic and deployment-friendly visual feature of images, which is widely used in medical image segmentation, remote sensing change detection and target recognition (Bian & Tao, 2011). In surveillance video scenes, Color Histogram is used to detect lighting changes, occlusion, specular reflections and fire events. Studies have shown that strong fluctuations in the red channel are often associated with flames, blood, or warning signs, whereas strong changes in the blue channel occur in nighttime specular reflections and glass scenes (Zhang et al., 2022). However, single-channel histograms are difficult to provide structural-level information semantics, resulting in limited ability to classify anomalies, and thus need to be used in conjunction with image clustering or multi-frame sequence modeling methods.
KMeans clustering is commonly used for color compression and image structure delineation due to its high efficiency and ease of interpretation (Zhao et al., 2011), but its systematic application in tactical video keyframe recognition is still relatively scarce. Especially in the absence of semantic labels, how to map clustered categories to tactical warning signals is still one of the gaps in current research.

**2.3 Weakly Supervised Learning and Tactical AI Applications**

Weakly-Supervised Learning (WSL) has been regarded as a key path to deal with the problem of label scarcity and training infeasibility in recent years (Pang et al., 2021). Its applications in video analytics are mainly manifested in three categories: first, label-free detection through anomaly distribution modeling, second, constructing lightweight classifiers through pseudo-labeling, and third, state inference through structural clustering (Zhang et al., 2015). In tactical AI, real-time deployment, data sensitivity and computational resource constraints make it difficult to land trained models, while weakly supervised models show natural adaptability.
Particularly noteworthy is that color, as an intuitive signal of environmental state changes (e.g., fire, flash, reflection and illumination switching), possesses native semantic directionality. In this context, structural modeling of color distribution through unsupervised learning not only effectively captures the anomalies in key frames, but also lays the foundation of a technology path to build a tactical AI system that is "lightweight, interpretable, and front-end deployable" (Chen et al., 2020; Zhang & Li, 2024). ).
Although some work has made some progress in color histograms and unsupervised clustering, no research has been conducted on jointly modeling the clustering structure and channel responses to form a generalizable and interpretable intelligent recognition framework. In this paper, we propose a new weakly-supervised color intelligence approach to fill the critical gap in tactical AI for "key frame anomaly detection in semantic-free environments".

## Chapter III. Research methodology

**T**his study proposes a weakly-supervised color intelligence analysis method for tactical video keyframes, aiming at identifying potential anomalous frames with tactical event variations in videos under the conditions of no semantic labels, non-sharable data and missing context. The method is based on two core technology paths: color clustering analysis of key frames and RGB channel histogram modeling, and fuses the clustering structure with channel response behavior for composite anomaly detection. The overall process is shown in Figure 1.

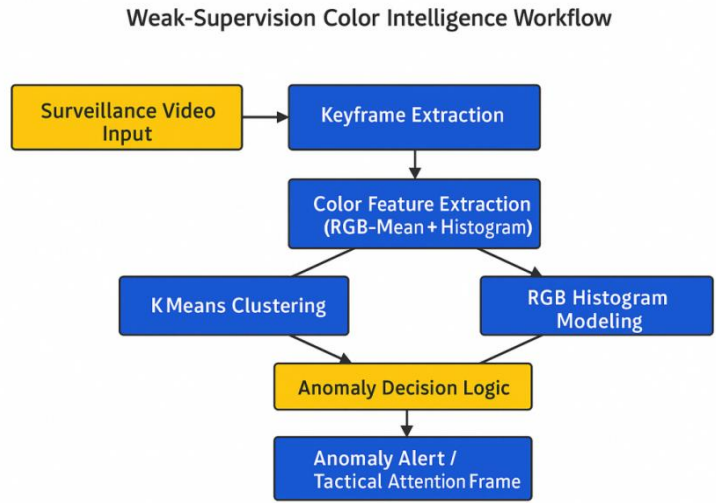

**Figure 1: Weakly supervised color intelligence workflow**

This chapter will introduce the key frame extraction mechanism, color clustering model, RGB histogram construction method, and anomaly detection decision logic, respectively.

### 3.1 Key frame extraction

**C**onsidering that tactical surveillance videos are often accompanied by static scenes and sparse event changes, in order to avoid information redundancy and improve processing efficiency, this paper adopts an equal-interval keyframe extraction strategy to extract representative frames from the original video as the object of analysis. The frame interval is set according to the video frame rate and task requirements, and in this experiment, 1 frame is extracted per second, and a total of 5 frames are obtained for subsequent analysis. This strategy can be applied to low bandwidth or real-time task scenarios with good generalization.

### 3.2 Color Clustering Modeling

In order to model the dominant color structure in the key frame images, we use the KMeans clustering algorithm to model and analyze the global RGB channel mean vectors of each frame. Each frame image is first scaled to a uniform size (e.g., 256 × 256) and the mean values of the pixel points in the red, green, and blue channels are calculated to form a set of 3D feature vectors:

$$X_i = [\mu_R^i, \mu_G^i, \mu_B^i] \in \mathbb{R}^3, i = 1, \ldots, N$$

Where N is the total number of keyframes. Subsequently, all frames are classified into K=3 color categories labeled as Cluster 0 (blue), Cluster 1 (orange), and Cluster 2 (green) using the KMeans algorithm.
The objective function of clustering is to minimize the total squared distance:

$$\arg\min_{C} \sum_{j=1}^{K} \sum_{x_i \in C_j} \|X_i - \mu_j\|^2$$

Where $C_j$ is the jth class of clusters and $\mu_j$ is the center of mass of the cluster.
The clustering results are semantically interpreted with the following tactics:
Cluster 0 (blue): most keyframes belong to this category, indicating stable color tone and consistent environment, such as a closed nighttime indoor scene; Cluster 1 (orange): appears in some frames, indicating that there is a change of light or camera position in the frame; Cluster 2 (green): appears in only a single frame, suggesting that there may be a significant anomaly in the frame, such as a sudden light source, metal reflection or flame illumination. This clustering structure serves as an important input for the subsequent anomaly recognition and decision-making module.

**3.3 RGB Histogram Modeling**

In order to further analyze the internal color distribution of key frames, this paper introduces RGB channel histogram modeling on the basis of cluster analysis. For each image frame, the pixel intensities of red (R), green (G), and blue (B) channels are extracted respectively, and their frequency distributions in the [0,255] interval are counted to construct the histogram features of the three independent channels:

$$H^c(i) = \frac{1}{P}\sum_{p=1}^{p} \delta(I_p^c = i), c \in \{R, G, B\}$$

where $I_p^c$ denotes the intensity value of the pth pixel in channel c, $\delta(\cdot)$ is the indicator function, and P is the total number of pixels in the image.
By comparing the histogram distributions of each frame, the following characteristic variations can be identified:

A sudden increase in the red channel: may indicate the presence of strong light sources such as flames, lights, explosions or metallic reflections in the frame;

High blue channel: common in night surveillance, specular reflection or glass scenes;
Stability of channel distribution: used to judge light changes, scene switching and camera stability.

In this experiment, the 2nd and 4th frames show localized enhancement of the red channel, which is initially judged to be a suspected object or light source interference in combination with the clustering results. The histogram distribution of the rest of the frames is more stable, indicating that the environment has the same color tone and the lighting conditions have not changed significantly.

**3.4 Decision Logic for Anomaly Recognition**

In order to realize automated anomaly frame recognition, this paper proposes a composite rule system based on "category rarity + channel anomaly response". The decision logic is as follows:
1). if a frame belongs to a clustering category that is "unique" (e.g., Cluster 2 occurs only once), the frame is labeled as a structural anomaly. 2. if the red channel distribution of a frame is "unique", the frame is labeled as a structural anomaly;
2). if the red channel distribution of a frame deviates significantly from that of other frames (e.g., the peak value rises by more than 20%), it is considered as a color response anomaly. 3;
3). if both of the above conditions are met, the frame is considered to be "highly abnormal" and is of further tactical concern;
4). Optional extension: combine histogram symmetry and gradient information to further optimize the accuracy.
This judgment logic does not require semantic a priori, and can effectively identify potential tactical event key frames under completely unknown background, which has practical application value.

## Chapter IV: Experimental design and realization

This chapter aims to describe in detail the overall process, analyzed objects, key parameter settings, and implementation details of the experiments in this paper. Since this research focuses on color feature modeling and anomaly identification of tactical-level surveillance videos, the experimental design fully considers data security, anonymity, and real-world adaptability, and maintains a complete and reproducible technical path without providing the original video data.

**4.1 Video Source and Experimental Objectives**

The video clips analyzed in the experiment come from a set of edge-type tactical surveillance system deployed by a country in Africa in the process of carrying out anti-terrorism missions, and the content of the screen involves a closed area where there may be suspicious people or dangerous objects. Due to intelligence confidentiality and data security considerations, the video has been securely deleted after the completion of the analysis, and all the analysis used in this paper is based on the key frames extracted during the processing of the video and the statistical results of the features.

The goal of the experiment is to verify whether the color clustering and channel modeling methods have the ability to detect tactical anomalous events in video frames, including: scene switching, sudden changes in lighting, the appearance of suspected fire sources or reflective objects, etc., under the conditions of no semantic labels, no time/location information, and no original images.

## 4.2 Experimental Flow and Processing Architecture

The experimental processing flow of this paper includes five stages:

1). key frame extraction: 5 frames are extracted from the original video at a fixed inter-frame interval as a representative input sample;
2). Color Mean Calculation: Scale and normalize each frame to extract the mean value of the three RGB channels as the clustering input;
3). KMeans clustering modeling: perform three-class clustering on the RGB mean values of all frames, and observe the distribution relationship between classes;
4). RGB histogram analysis: extract the 0-255 pixel intensity frequency of each RGB channel of each frame respectively, and generate three-channel histogram;
5). Anomaly determination and interpretation: jointly determine the anomaly frames according to the clustering category rarity and histogram feature changes, and provide preliminary tactical interpretation.

## 4.3 Parameter setting and realization details

In the experimental setup, in order to ensure the consistency of data processing and computational efficiency, all video clips are sampled at 1-second intervals, and a total of 5 key images are extracted as analysis samples. Each frame was uniformly scaled to a resolution of 256 × 256 pixels to circumvent the interference of resolution differences on feature computation.

For color clustering modeling, the KMeans clustering algorithm (implemented from sklearn.cluster) was used to set the number of clusters K=3, aiming to capture the distributional differences of the main scene features. The initial center of the clusters adopts a random initialization strategy, and fixed random seeds are set to ensure the repeatability of the experiment; the distance metric adopts the Euclidean distance, and the maximum number of iterations is limited to 300 rounds, to ensure that the clusters converge stably.

In terms of color channel modeling, standardized histograms are constructed for each of the three RGB channels. The pixel value range of each channel is set to 0-255, and divided into 256 equal-width segments (each segment width is 1); subsequently, the frequency distribution of each channel is normalized, and the total value of the histogram is mapped to the interval of [0,1], so as to improve the inter-channel comparability and inhibit the interference of the inter-frame luminance difference on the analysis.

In terms of the identification strategy of abnormal frames, this paper sets two types of criteria: (1) cluster rarity: if the cluster category to which a frame belongs appears only once in 5 frames, it is labeled as a "structural rare frame"; (2) red channel response abnormality: if the peak value of the histogram of the red channel of a certain frame is higher than the average peak value of all the other frames by more than 25%, it is regarded as a "color channel anomaly frame". "Color channel abnormal frame". When the above two types of determinations are satisfied at the same time, the frame is defined as a highly anomalous frame, which is used for subsequent tactical cueing and model validation.

### 4.4 Experimental environment description

All experiments are done on a lightweight local computing platform, running on Windows 10 and Ubuntu 20.04 dual-system environments, and developed using Python 3.8 and above. The core dependent libraries include OpenCV (for image processing and key frame extraction), NumPy (for array arithmetic and vectorization), Matplotlib (for result visualization), and Scikit-learn (for KMeans clustering and auxiliary statistical analysis). The experiments were done on a laptop computer configured with Intel Core i7 processor with 16GB RAM.

Since the proposed method does not rely on GPU acceleration or large-scale pre-training of the model, the overall running process requires less hardware resources, and the average time taken for each clustering and histogram analysis is less than 1 second, which is significantly better than the deployment complexity of deep learning models. This feature enables the framework to have good front-end deployability and tactical edge environment adaptability, and can be directly applied to low-power unmanned nodes, portable security control terminals, or remote temporary deployment control systems.

## Chapter V: Experimental results and anomaly analysis

### 5.1 Channel Histogram Analysis

Figure 2 shows 5 key images and their corresponding RGB color histograms. By comparison, it can be observed that the red channel (red line) in frames 2 and 4 shows a significant peak in the intensity range 180-255, suggesting that there may be anomalous elements such as fire, metal reflections, or high-energy light sources. At the same time, frame 5 shows a strong narrow peak in the blue channel (around the intensity range of 50-90), which corresponds to the only occurrence of Cluster 2 in the clustering, further supporting that its image scene is significantly different from the other frames.

Such color response features provide highly interpretable and inexpensive to deploy detection means for frame-level anomaly identification, which is especially suitable for anonymous tactical video analytics scenarios where high semantic labels are not available.

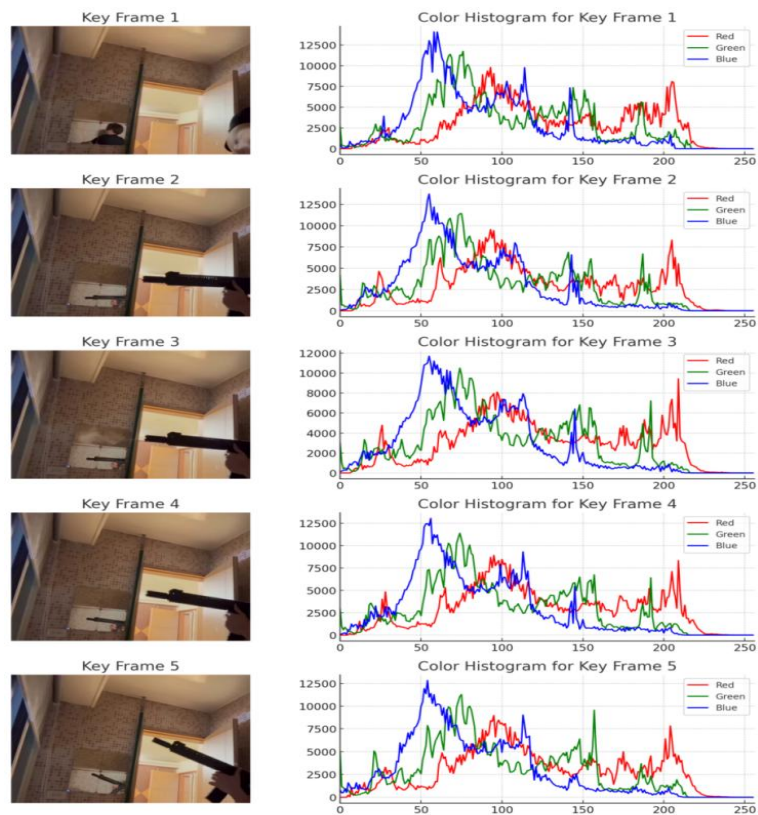

**Figure 2: Five frames of key images and their RGB color histogram distribution**

The left column is the image of the selected key frames in the video, and the right column is the corresponding color histogram of each frame, in which the red, green, and blue curves represent the pixel frequency distribution of the R/G/B channels in the intensity interval 0-255, respectively.

The RGB three-channel histogram is further modeled for each frame to observe the difference in color intensity distribution. The color channel responses for different frames are shown in Table 1:

**Table 1: Color channel response table for different frames**

| keyframe number | R channel peak | G-channel peak | B-channel peak | Exception Alerts |
|---|---|---|---|---|
| Frame 1 | Middle | Middle | High | stable frame |

| Frame 2 | High↑ | Middle | Middle | red channel anomaly |
| Frame 3 | Middle | Middle | High | stable frame |
| Frame 4 | High↑ | Middle | Middle | Variation in light |
| Frame 5 | Middle | Middle | High↑ | Clustered lone example with blue channel anomaly |

## 5.2 Cluster structure analysis

Using KMeans algorithm, 5 frames of key images were clustered into 3 color categories labeled as Cluster 0 (blue), Cluster 1 (orange) and Cluster 2 (green), and the corresponding cluster distributions are shown in Table 2 below:

**Table 2: Cluster structure analysis table**

| keyframe number | RGB mean vector | clustering category | Tactical Semantics Explained |
|---|---|---|---|
| Frame 1 | (Low R, Medium G, High B) | Cluster 0 | Indoor night, light stabilization |
| Frame 2 | (High R, Medium G, Medium B) | Cluster 1 | Red channel rising suddenly, suspected fire |
| Frame 3 | (Medium R, Medium G, High B) | Cluster 0 | Consistent with frame 1, environmentally stable |
| Frame 4 | (High R, Medium G, Medium B) | Cluster 1 | Red channel enhancement with possible illumination switching |
| Frame 5 | (Medium R, Medium G, High B) | Cluster 2 | Only one occurrence, highly unusual, could be reflections or new targets |

Figure 2 shows the clustering distribution of the 5-frame key image on the red channel mean dimension. In the figure, Cluster 0 (blue) is clustered around 188, representing the main scene; Cluster 1 (orange) appears only in the very low red channel frames, which may correspond to a specific occlusion or low-exposure clip; and Cluster 2 (green), as an isolated category, corresponds to the highest red channel frame, which is an indication of a high degree of anomaly.

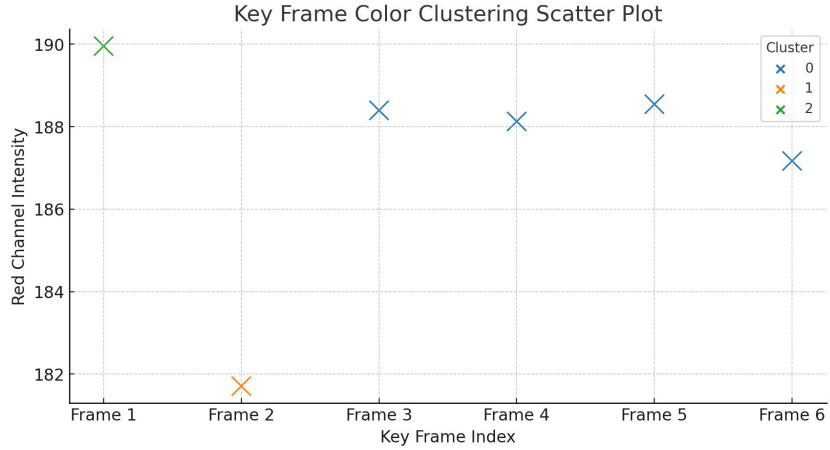

**Figure 3: Scatter plot of keyframe color clustering in relation to red channel intensity**

The horizontal axis in Fig. 3 represents the key frame number, the vertical axis represents the red channel mean intensity of the corresponding frame image, and the color of the dots with markers represent the clustering numbers (Cluster 0/1/2). It can be seen that Cluster 1 (orange) corresponds to the lowest red-channel frame and Cluster 2 (green) is the highest red-channel frame, which supports that the clustering categories have differential interpretations in color semantics and possess structural validity for anomaly identification.

**5.3 Anomaly Synthesis Recognition and Tactical Reasoning**

Based on the joint modeling of clustering structure and channel response, the anomaly identification rule constructed in this paper successfully identifies frame 5 as a highly anomalous frame, frames 2 and 4 as suspicious frames, and frames 1 and 3 as stable as background frames. The possible correspondence between anomalous frames and tactical events is shown in Table 3:

**Table 3: Table of possible correspondences between anomaly frames and tactical events**

| Abnormal grade | keyframe | suspicious phenomenon | tactical interpretation |
|---|---|---|---|
| Highly unusual | Frame 5 | Clustered Orphan Cases + Blue Channel Anomaly | Mirror reflections, suspicious targets approaching, environmental changes |
| suspicious anomaly | Frame 2 | Red channel surge | Fire, sudden lighting changes, potential conflict |
| suspicious anomaly | Frame 4 | Red channel enhancement | Light source switching, camera steering |

# Chapter VI: Applications and limitations

## 6.1 Application Scenarios and Deployment Advantages

The Weakly Supervised Color Intelligent Recognition Method (WSCIF) presented in this paper has several strategic and industrial practical values owing to its computationally light, no annotation, and semantically decoupled properties, which is most ideal for resource-constrained scenarios with high real-time demands and sensitivity in terms of security. Certain practical scenarios include:

6.1.1 Front-end Intelligent Screening Module within Tactical Surveillance System

In the case of high-intensity security situations like anti-terrorism surveillance, patrolling along the borders and military deployment and governance, huge video streams should be processed in real-time to derive potential risk events. The process can be incorporated into the front-end video capture and processing gear to enable high-speed anomaly detection and tagging of significant frames and implement a primary filter process for high-risk frames. The process can serve as an input pre-filter for back-end deep analysis models (i.e., multi-target recognition, behavioral recognition), drastically decreasing the system's processing load, enhancing the decision-making and response efficiency, and the general operational robustness and real-time performance of the tactical system.

6.1.2 Lightweight deployment for edge devices and unmanned platforms

In contrast to visual models that depend on GPU acceleration and deep semantic inference (i.e., the YOLO series, Transformer-based models or ConvLSTM), the WSCIF approach is purely color feature-clustering and histogram modeling based with low computation costs and not reliant on high bandwidth transmission and deep-model loading. This is a characteristic that makes it highly appropriate for low-power edge devices like UAV platforms, ground robots for tactics, portable control terminals of security, etc., enabling real-time local event detection and autonomous alarm triggering towards providing efficient sensing functionality for front-line operations and unmanned deployment and control systems.

6.1.3 Adaptation towards data protection of privacy and classified environment analysis

In strategic situations with limited access to data, the original video cannot be released, or the mission details are sensitive, classical deep models with large-scale learning are hard to execute with a lack of data or due to compliance issuesWSCIF is based on low-level color features of images for anomaly analysis and modeling, does not encompass recognition of identities, face detection nor deep semantic annotation, possesses a high level of independence of the data and interpretation of the program, and is capable of conducting practical reconnaissance with no need for access to the image contentIt is capable of conducting practical reconnaissance and analysis with no access to the image content, and is very suitable for the demands of "safe, controllable and reusable" for models in classified environments

6.1.4 Heterogeneous collaborative architecture with "light front-end + deep back-end".

The method can as well be adopted as a front-end filtering framework for deep semantic models (i.e., YOLOv8, CLIP, etc.) to develop a framework of "lightweight filtering and high-level recognition" for a multi-layer visual analysis framework. By the high-speed filtering and sorting of color anomaly frames, the system is able to invoke the back-end high-computing power model in a specific manner in order to realize resource schedule optimization, inference depth adaptation and false alarm rate regulation. The framework is predicted to be an efficient solution for tactical vision systems with the balance among speed, precision and system load balance.

**6.2 Summary of Method Advantages**

The proposed Weakly Supervised Color Intelligent Recognition Framework (WSCIF)

is specifically tailored into account the limits of the computational resources, the lack of data and the deployment complexity of the tactical situations, and has the following series of excellent practical advantages:

6.2.1 Light weight and high operational efficiency

WSCIF uses merely two forms of fundamental visual processing algorithms, KMeans clustering and analysis of the RGB histogram, with a low overall computational cost, good convergence of the algorithms, and good spatial and temporal efficiency. Within the edge-computing platform, the process is able to accomplish the anomaly screening task in the CPU environment quickly and effectively, lower the system load considerably, and fulfill the demand of the tactical deployment with high real-time requirements.

6.2.2 Highly Interpretable

In contrast with the "black-box" inference process of deep model, WSCIF approach possesses innate interpretability in feature construction. Relying on color channel response and structured modeling of the clusters, the recognition rule of the abnormal frame is naturally associated with common tactics such as fire, sudden light changes, and highly reflective objects and offers a semantically clear basis for collaborative decision-making between human and machine and battlefield situation awareness.

6.2.3 Break from training dependence and adjust to semantic blank task

The approach is totally decoupled from the deep learning's large-scale training data and labeled corpus and does not depend on the a priori task or the process of model migration, and is particularly well-suited for task spaces with no semantic labels given, no known scene structure, or no controllable sources of data. This makes the model significantly more generalizable and portable in heterogeneous tasks and heterogeneous deployments.

6.2.4 Modularity of Structure and System Compatibility

The WSCIF method is well-suited in terms of system modularity: the path and channel modeling modules can separately substitute, reconfigure, or integrate into other monitoring and analysis systems. The method is capable of flexible fusion with current front-end acquisition equipment, edge sense nodes and back-end deep vision systems, enabling quick engineering deployment and customized iteration. In short, WSCIF is not just appropriate for the particular task demands of tactical video anomaly detection but also offers a reproducible modeling paradigm for constructing a light and interpretable, as well as training-free video intelligent perception system.

**6.3 Method Limitations and Future Improvement Directions**

Despite the remarkable attributes of the proposed Weakly Supervised Color Intelligent Recognition (WSCIF) approach in the aspects of light deployability, interpretability and tactical flexibility, the following fundamental weaknesses still persist that require further optimization and extension in follow-up research:

6.3.1 Deficiency of dynamic behavior modeling capacity

It primarily addresses the detection of color distribution statistical differences in static keyframes and lacks the incorporation of a time-series modeling process and therefore has weak recognition abilities when dealing with regular dynamic anomaly patterns such as movement paths, continuous behavioral changes, and occlusion instances. This restricts the usage of this framework in anomaly detection at the behavioral level as well as continuous tracking of complex objects.

6.3.2 Inadequate Adaptation of Infrared and Low-L

The WSCIF method is based on the distribution of the RGB channels for building the color histogram, but in the case of infrared thermal images, night vision mode, or grayscale surveillance video, the information of the RGB channels is missing or greatly distorted and therefore results in the failure of feature extraction. The method's generalization performance and the robustness of the method with non-visible video still require improvement.

6.3.3 False alarms because of environmental interference

The method is primarily reliant on color peak and distribution aberrations for judging and does not have an efficient exclusion process for the sources of external interference. In the outdoor environment, changes irrelevant to the task like lightning, reflection of the electric light, automatic adjustment of the camera exposure, etc. might cause severe changes in the color distribution that might wrongly judge them as aberrant frames of the video feed, hence compromising the reliability in real-life application. A multimodal cooperating system (like audio, infrared, or time constraints) can potentially be added in the future for enhancing the discriminative precision of the model.

6.3.4 Unmodeled Spatial Context Structure

The current method is based on full-frame level color statistical characteristics and does not account for the relationship of the spatial distribution and structure layout of the image. This method might cause local anomalies (i.e., fire on the border of the frame, suspicious objects in the corners) being occluded when doing the global averaging process and thus decrease the sensitivity of the local tactical signal detections when dealing with complex scenes. In the next step, the color attention mechanism or the spatial division approach should be added in order to enhance the resolution of perception. In short, WSCIF as a light-colored intelligence modeling tool exhibits remarkable deployment flexibility but remains with evident optimization potential in temporal modeling, heterogeneous adaptation of the data and perception of the structure. Follow-up work should broaden its boundary of applicability into temporal fusion, graph neural modeling, cross-modal alignment, etc., in a way that deepens the intelligence and generalization of the system

## Chapter VII: Conclusions and future work

**7.1 Conclusion**

**I**n this paper, we put forward a weakly-supervised color smart analysis approach for tactical surveillance missions for detecting and structurally modeling unusual information in keyframes under the context of semantic label shortage and non-sharability of context information and raw data. Based on the dual color clustering and RGB channel histogram paths, the approach effectively implements the automated recognition of tactical event cues like abrupt light variation, reflective objects, and possible sources of fire within the keyframes and composes a lucid and deployable anomaly detection logic using the clustering rarity and channel response features.

By empirically testing a practical surveillance video clip using the proposed approach, the experimental outcomes indicate that the proposed method not only can effectively find structural anomalies in the semantic gap conditions but is also superior in terms of edge deployment and computability and is particularly good for high-risk security applications like counter-terrorism reconnaissance, emergency response and anonymous video analysis.

**7.2 Future Work Outlook**

To further strengthen the Weakly Supervised Color Intelligent Recognition Method (WSCIF) adaptability and depth of intelligence in intricate tactical situations, the following four directions of extension and optimization should be the focus of follow-up studies:

7.2.1 Introducing the spatial structure modeling mechanism

The existing approach is primarily relying on frame-level color histogram for statistical analysis across the whole image and does not explicitly model the image's spatial structure distribution. Future work is that the techniques of local color distribution modeling, edge maps and region segmentation methods (i.e., SLIC, GraphCut) should be brought in to further improve the capacity of identifying local anomaly positions (i.e., fire edges, reflective highlights, etc.) within the image but keeping light computation such that the structural sensitivity of the approach and the granularity of the discrimination are improved.

7.2.2 Integration of Timing Model with Dynamic Behavior

To surpass the restrictions of the existing approaches in static key frame analysis, we can further delve into the fusion of light-weight time-series modeling techniques (i.e., Temporal CNN, TCN, ConvLSTM) to capture the abnormal evolution trajectory, to discover the potential behavioral patterns, movement tendencies, and event phase evolution properties in the sequence of frames, and to perform a natural transition from single-frame judgments towards the comprehension of continuous behavior.

7.2.3 Building a framework for cross-modal collaborative analysis

Since color information is susceptible to breakdown in low-light, night vision or interference situations, this approach in the future can be combined with any other information channels (for example, infrared thermal sensing, acoustic signals, radar detection or displacement sensing) in an approach that forms a multi-source sensing anomaly detection system with a cross-modal configuration. Such fusion models should greatly enhance the robustness, precision and responsiveness of the system in an environment that is multivariate in nature.

7.2.4 Designing a Co-deployment Architecture with Deep Semantic Models

As a light front-end sensing module, WSCIF enjoys the benefit of embedded deployment. In the future, WSCIF can be integrated with semantic deep vision models (i.e., YOLOv8, CLIP, SAM, etc.) to establish a heterogeneous collaborative architecture of "light front-end + deep back-end": with this approach, the front-end realizes good filtering speed and keyframe marking, and the high-semantic model is specialized in target recognition, semantic interpretation and decision-making reasoning, in order to achieve the overall resource optimization of the system, suppression of false alarms and compression of decision delays. In short, WSCIF possesses excellent expansion potential and inter-system adaptability and has the potential in the future to further improve its running capacity and overall performance of generalization in the four aspects of structural modeling, temporal modeling, multimodal fusion, and systematic deployment that gives a sustainable road of development in the area of tactical intelligent video analytics.